\documentclass[11pt]{article}

\usepackage[final]{acl}

\usepackage{times}
\usepackage{latexsym}

\usepackage[T1]{fontenc}

\usepackage[utf8]{inputenc}

\usepackage{microtype}

\usepackage{inconsolata}

\usepackage{graphicx}

\usepackage{booktabs}   
\usepackage{array}      
\usepackage{amsmath}
\usepackage{amssymb}
\usepackage{listings}
\usepackage{xcolor}
\usepackage{algorithm}
\usepackage{algpseudocode}

\title{MM-Doc-R1: Training Agents for Long Document Visual Question Answering through Multi-turn Reinforcement Learning}

\author{
 \textbf{Jiahang Lin\textsuperscript{1}$^\ddagger$}\thanks{Equal contributions.\ \ $^\ddagger$Work done during an internship at Shanghai Qiji Zhifeng Co., Ltd.\ \ $^\dagger$Corresponding authors: \href{mailto:hzhua201@gmail.com}{hzhua201@gmail.com}, \href{mailto:tgui@fudan.edu.cn}{tgui@fudan.edu.cn}.},
 \textbf{Kai Hu\textsuperscript{2}$^*$},
 \textbf{Binghai Wang\textsuperscript{1}},
 \textbf{Yuhao Zhou\textsuperscript{1}},
 \textbf{Zhiheng Xi\textsuperscript{1}},
\\
 \textbf{Honglin Guo\textsuperscript{1}},
 \textbf{Shichun Liu\textsuperscript{1}},
 \textbf{Junzhe Wang\textsuperscript{1}},
 \textbf{Shihan Dou\textsuperscript{1}},
 \textbf{Enyu Zhou\textsuperscript{1}},
\\
 \textbf{Hang Yan\textsuperscript{2}},
 \textbf{Zhenhua Han\textsuperscript{2}$^\dagger$},
 \textbf{Tao Gui\textsuperscript{1}$^\dagger$},
 \textbf{Qi Zhang\textsuperscript{1}},
 \textbf{Xuanjing Huang\textsuperscript{1}}
\\
 \textsuperscript{1}Fudan University,\ \
 \textsuperscript{2}Shanghai Qiji Zhifeng Co., Ltd
\\
}

\begin{document}
\maketitle
\begin{abstract}
Conventional Retrieval-Augmented Generation (RAG) systems often struggle with complex multi-hop queries over long documents due to their single-pass retrieval. We introduce \textbf{MM-Doc-R1}, a novel framework that employs an agentic, vision-aware workflow to address long document visual question answering through iterative information discovery and synthesis. To incentivize the information seeking capabilities of our agents, we propose \textbf{Similarity-based Policy Optimization (SPO)}, addressing baseline estimation bias in existing multi-turn reinforcement learning (RL) algorithms like GRPO. Our core insight is that in multi-turn RL, the more semantically similar two trajectories are, the more accurate their  shared baseline estimation becomes. Leveraging this, SPO calculates a more precise baseline by similarity-weighted averaging of rewards across multiple trajectories, unlike GRPO which inappropriately applies the initial state's baseline to all intermediate states. This provides a more stable and accurate learning signal for our agents, leading to superior training performance that surpasses GRPO. Our experiments on the MMLongbench-Doc benchmark show that \textbf{MM-Doc-R1} outperforms previous baselines by \textbf{10.4\%}. Furthermore, \textbf{SPO} demonstrates superior performance over \textbf{GRPO}, boosting results by \textbf{5.0\%} with Qwen3-8B and \textbf{6.1\%} with Qwen3-4B. These results highlight the effectiveness of our integrated framework and novel training algorithm in advancing the state-of-the-art for complex, long-document visual question answering.
\end{abstract}

\section{Introduction}
Long document visual question answering presents a challenging yet highly practical research problem, primarily due to the difficulty of effectively identifying and extracting salient information from lengthy, multi-page documents~\cite{van2023document,appalaraju2021docformer,ma2024mmlongbench}. Existing work is always based on Retrieval-Augmented Generation (RAG), where textual or visual content is encoded into embeddings, and relevance is determined by similarity scores with respect to the original query~\cite{peng2024graph,lewis2020retrieval,han2025mdocagent}. 
These approaches typically rely solely on the initial user query for retrieval, which limits their effectiveness in handling multi-hop questions that require iterative information gathering across multiple document segments.

\begin{figure}[t!] 
    \centering 
    \includegraphics[width=\linewidth]{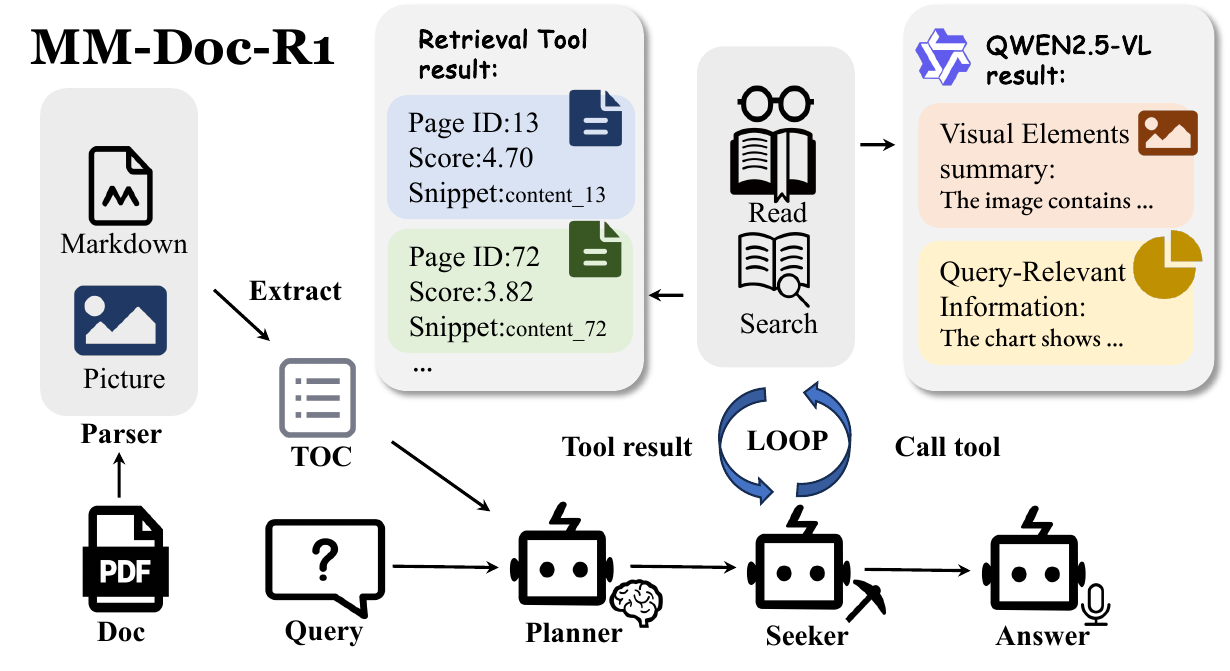} 
    \caption{Introduction to MM-Doc-R1. MM-Doc-R1 employs a seeker for iterative key information retrieval within documents, leveraging a VLM (Visual Language Model) as a reading tool to ensure accurate processing of visual elements.} 
    \label{fig:introduction} 
\end{figure}

To address the limitations of existing models in multi-turn information retrieval from documents, we propose \textbf{MM-Doc-R1}, a novel framework that integrates a vision-aware long document question answering workflow with an end-to-end multi-turn reinforcement learning algorithm. As illustrated in Figure~\ref{fig:introduction}, our workflow comprises three specialized agents. First, a planner generates an initial information-seeking plan by parsing the document's table of contents. Following this, a seeker acts as a tool-driven agent, performing multi-turn information retrieval through iterative interactions with the document to gather relevant evidence. The seeker utilizes a ``search'' tool for text-based retrieval and a ``read'' tool, powered by a vision-language model (VLM), to extract visual details from specific pages. This iterative decomposition of sub-questions and strategic tool invocation enables the seeker to precisely access relevant pages and retrieve accurate information. Finally, the collated relevant information is fed into an answer agent to generate the ultimate response. To enhance the agents' information-seeking capabilities and refine their decision-making throughout this iterative workflow, we employ \textbf{Multi-turn Reinforcement Learning} to train our agents.

In multi-turn reinforcement learning, GRPO~\cite{song2025r1,song2025yingqian} is commonly employed. It operates by first generating a complete trajectory through rollout and then computing the advantage as the difference between the total accumulated reward and a baseline. However, GRPO estimates this baseline only from the initial state's rollout, which is then inappropriately applied to intermediate states. This introduces significant estimation bias in those intermediate steps.
To tackle this problem, we are introducing SPO, a new multi-turn RL algorithm built on GRPO. Our core idea is this: the more semantically similar two trajectories are, the greater their overlap in intermediate states. This increased overlap directly leads to a more accurate baseline estimation. SPO capitalizes on this by weighting rewards based on trajectory similarity, yielding a more accurate and consistent baseline estimation. This enhanced baseline effectively reduces variance and guides the learning process toward more precise convergence, thereby significantly boosting the agent's information-seeking ability in multi-turn RL training.

Extensive experiments on the MMLongbench-doc benchmark demonstrate that our method, \textbf{MM-Doc-R1}, outperforms previous RAG baselines by \textbf{10.4\%}. Furthermore, our proposed \textbf{SPO} approach delivers substantial improvements over \textbf{GRPO}, yielding a \textbf{6.1\%} performance increase with Qwen3-4B and a \textbf{5.0\%} performance increase with Qwen3-8B. These results collectively underscore the superior capability of MM-Doc-R1 in handling complex long-document and visually-rich question answering tasks. Our key contributions are summarized as follows:
\begin{itemize}

    \item We propose \textbf{MM-Doc-R1}, a novel end-to-end framework for long document visual question answering that integrates a vision-aware workflow with multi-turn reinforcement learning. Our framework empowers agents with iterative information discovery and synthesis, which significantly boosts retrieval accuracy and ultimately improves answering precision in complex document understanding.

    \item We introduce \textbf{Similarity-based Policy Optimization (SPO)}, a new reinforcement learning algorithm specifically developed to enhance agents' information-seeking capabilities within our framework. This approach provides more stable and accurate baseline estimates for agents in multi-turn settings, thereby enabling faster learning convergence and improved overall performance.

    \item We validate the effectiveness of MM-Doc-R1 through extensive experiments. Our approach consistently improves the performance of Qwen3 models with 4B and 8B parameters across various subsets of MMLongBench-Doc, achieving overall superior results compared to existing baselines.
\end{itemize}

\begin{figure*}[t!] 
    \centering 
    \includegraphics[width=1.0\textwidth]{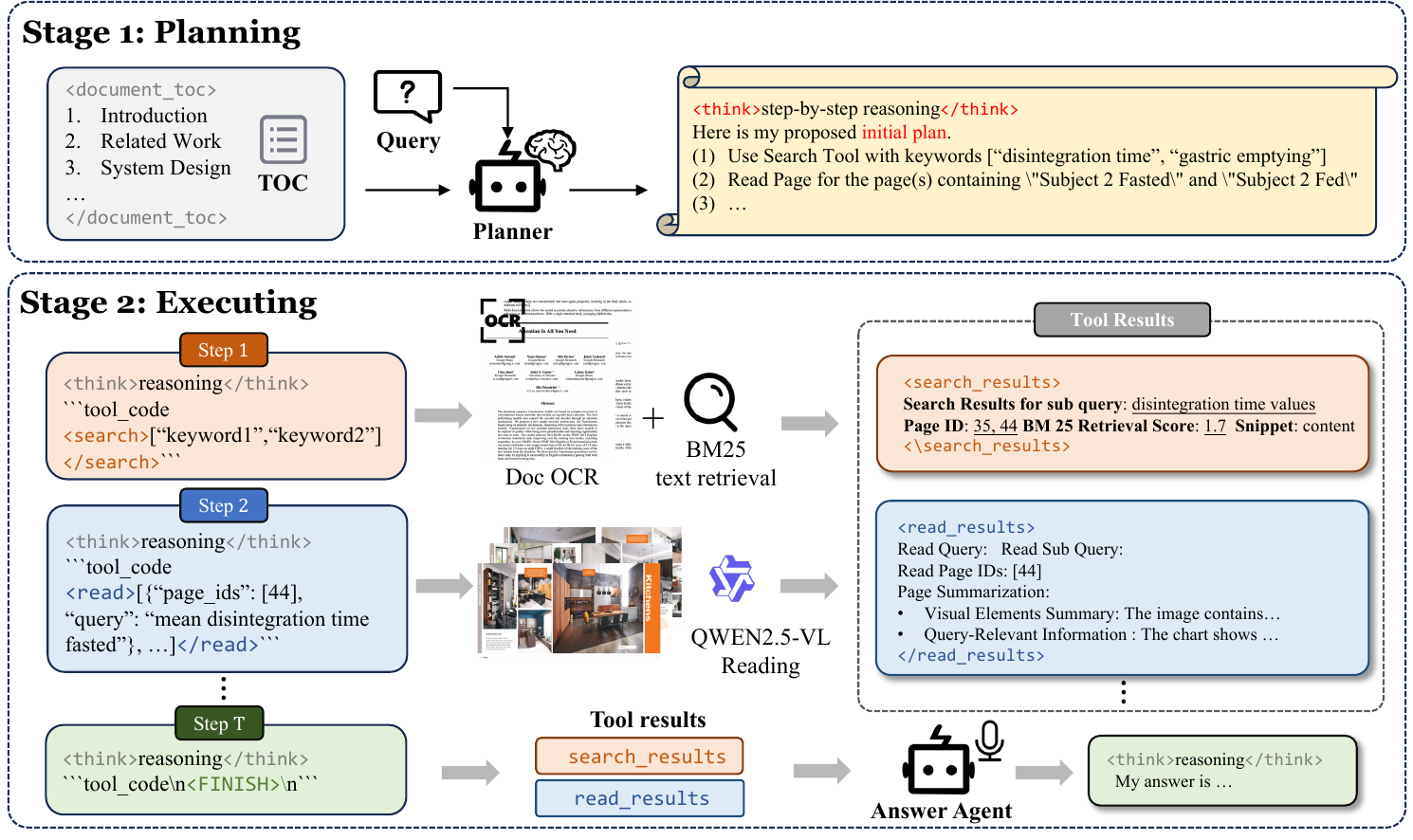} 
    \caption{Detailed framework of MM-Doc-R1.
The framework operates in three sequential stages. First, the planner module formulates a reasoning plan based on the preprocessed document TOC and the user query. Subsequently, the information seeker executes a multi-turn retrieval and reading process, leveraging the ``search" and ``read" tools to gather relevant information. Finally, the collected knowledge is integrated by the answer agent , which generates a coherent and contextually accurate response.} 
    \label{fig:main_figure} 
\end{figure*}
\section{Related Work}
\subsection{DocVQA Task}
The Document Visual Question Answering (DocVQA) task needs models to answer questions by jointly reasoning over both textual and visual information present in documents~\cite{mishra2019ocr,suri2024visdom,gao2023retrieval}. Various approaches extract visual information from figures as a means to process different modalities~\cite{memon2020handwritten,wu2024stateflow}. Early research primarily concentrated on extracting concise answers from single, short documents ~\cite{mathew2021docvqa}. However, with the advent of large language models, addressing QA tasks involving multiple or lengthy documents has become a significant challenge~\cite{yu2024visrag,tanaka2025vdocrag}. Some benchmarks like MMLongbench-Doc~\cite{ma2024mmlongbench} and LongDocURL~\cite{deng2024longdocurl} focus on the long document question answering.
One approach to tackle this involves employing Optical Character Recognition (OCR) and text-based retrieval to identify the most relevant document chunks~\cite{khattab2020colbert,zhang2024ocr}. Another method utilizes visual encoders to obtain visual embeddings, which are then used for retrieval based on a query's embedding; a recent example of this is ColPALI~\cite{faysse2024colpali}.
The embedding retrieval method proves to be a highly valuable tool, finding application in multi-agent systems. Consequently, several agentic approaches, such as M3docrag ~\cite{cho2024m3docrag} and MDocagent~\cite{han2025mdocagent}, have leveraged multi-agent systems to solve the DocVQA problem. These systems integrate text embedding-based RAG (Retrieval Augmented Generation) and image embedding-based RAG through different agents. By fostering cooperation between Vision-Language Models (VLMs) and Large Language Models (LLMs), these systems aim to achieve superior results in the DocVQA task.These methods primarily address single-hop questions. However, multi-hop questions necessitate a multi-turn approach and the generation of concise sub-queries. Our proposed method focuses on resolving this particular challenge.

\subsection{Retrieval-Augmented Generation}

RAG (Retrieval Augmented Generation) frameworks significantly enhance Large Language Models (LLMs) by integrating external knowledge retrieval~\cite{zhao2024retrieval,jiang2023active}, thereby enabling the generation of more factually grounded responses~\cite{han2023comprehensive}. While traditional retrieval methods, such as BM25 and dense retrievers like BGE-M3~\cite{chen2024bge}, excel at lexical or semantic matching, they often encounter limitations when tackling complex multi-hop queries~\cite{gao2023retrieval}.
Recent advancements have led to multi-modal extensions, exemplified by models like ColQwen2.5, which builds upon ColPALI~\cite{faysse2024colpali}, that incorporate visual features to enrich the retrieval process. However, these models still face challenges in terms of iterative refinement for complex question answering. Furthermore, some web search retrieval methods, such as Search-R1~\cite{jin2025search} and Deepreasearcher~\cite{zheng2025deepresearcher}, employ multiple retrieval steps to address long-context QA problems. Yet, these methods primarily focus on the text modality and rely on generic web search tools.
In contrast, our proposed method distinguishes itself by combining both visual and text modalities through the integration of a specialized ``visually-read" tool. This unique multimodal approach enables our method to address a broader range of problems within the visually-rich Question Answering (VQA) domain.

\subsection{Reinforcement Learning Algorithm}

The earliest and most widely adopted Reinforcement Learning method for training Large Language Models is Proximal Policy Optimization~\cite{schulman2017proximal}. PPO utilizes a critic model to estimate the baseline, which represents the average reward of all possible actions in a given state.
Recently, with the development of models like DeepSeek-R1~\cite{guo2025deepseek}, Group Relative Policy Optimization (GRPO,~\citet{shao2024deepseekmath}) is gaining increasing traction for training LLMs. Compared to PPO, GRPO estimates the baseline using a group mean reward, thereby eliminating the need for a separate critic model. This can lead to significant savings in computational resources and memory.
Other methods, such as REINFORCE++~\cite{hu2025reinforce++}, also propose alternative baseline estimations, like using the batch mean reward. More recently, VRPO~\cite{zhu2025vrporethinkingvaluemodeling} revisits value modeling for robust RL training under noisy supervision. This trend highlights ongoing research into more efficient and stable RL algorithms for LLM alignment.

\section{Method}

In this section, we present the core components of our proposed MM-Doc-R1 framework.
First, we design an autonomous, structured agentic workflow to flexibly process multi-page documents. This workflow consists of three key agents: a planner, a seeker, and an answer agent. This design allows our agents to iteratively search for crucial information within documents.
Secondly, we introduce an innovative reinforcement learning algorithm called \textbf{S}imilarity-based \textbf{P}olicy \textbf{O}ptimization (SPO). We use SPO to train our agents from scratch, and this training method significantly enhances our agents' information-seeking capabilities, empowering them to efficiently locate key information in documents.
Our workflow is illustrated in Figure~\ref{fig:main_figure}.

\subsection{Agentic Workflow}

To accurately and comprehensively respond to complex questions that necessitate integrating information from diverse sources or performing multi-step reasoning, our agent adheres to a meticulously structured workflow. This sequential yet iterative process not only mimics human cognitive approaches to problem-solving but also endows the model with dynamic planning and essential self-correction capabilities. The entire workflow unfolds across five distinct, interconnected phases.

\subsubsection{Document Parsing}
When a document is received, we first use the OCR tool Doc2X\footnote{\url{https://github.com/NoEdgeAI/pdfdeal}} to parse it, extracting tables, figures, and text into Markdown format. After we get the OCR output, we create a table of contents (TOC) by the markdown text, providing a main abstract of the document. Subsequently, the document is chunked by pages. This process generates both a TOC and a list of chunks. Each chunk is derived from an OCR result and includes a corresponding image, which is a screenshot of the relevant page.

\subsubsection{Initial Planning}
Following document parsing, a planning agent is employed to formulate a global strategy. Its inputs include the TOC and detected figures' caption. The planner is responsible for breaking down the initial query into granular sub-queries and orchestrating the selection of necessary tools. This global perspective, integrated into the seeker's history, critically informs and guides the agent's decision-making process.
\begin{figure*}[t!] 
    \centering 
    \includegraphics[width=0.85\textwidth]{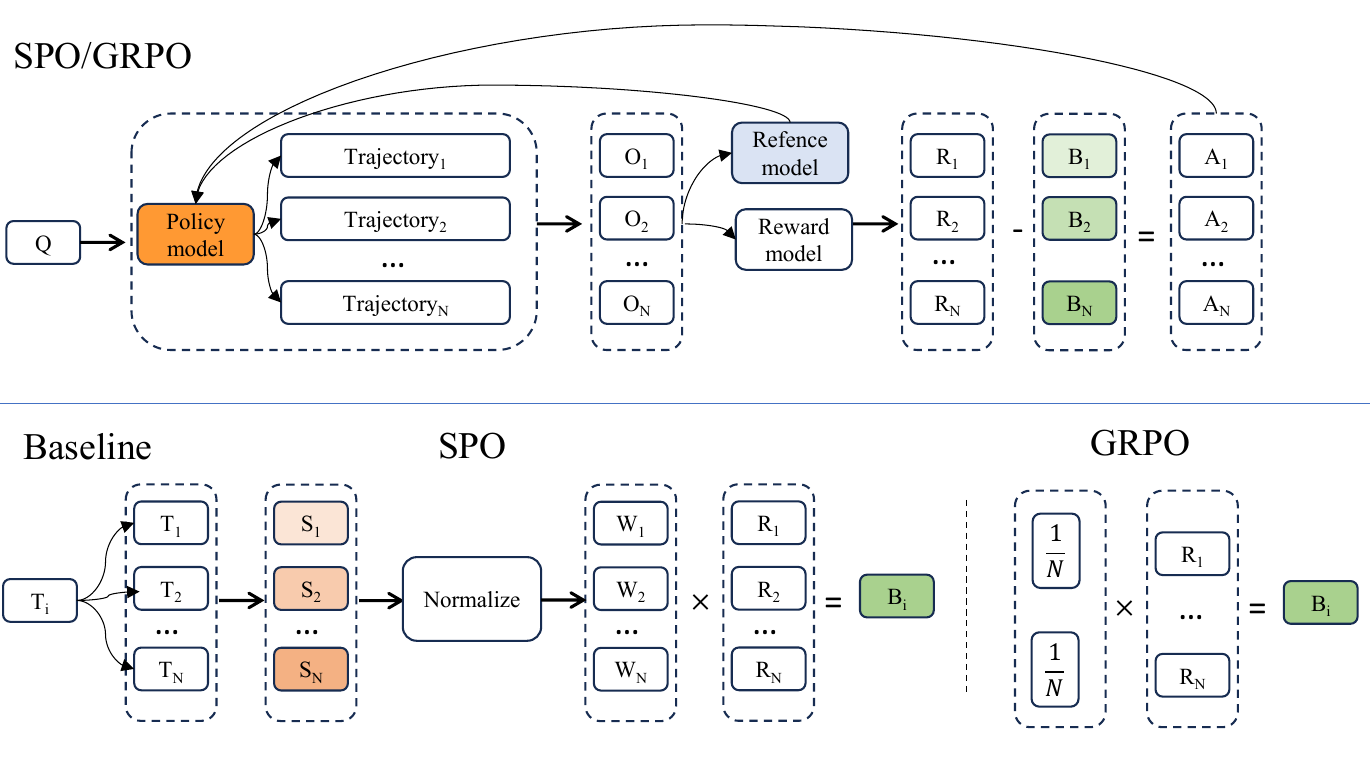} 
    \caption{SPO and GRPO's advantage estimation. The bottom panel shows the baseline computation of SPO and GRPO.} 
    \label{fig:algorithm1} 
\end{figure*}
\subsubsection{Toolbox}
Our agent is equipped with two specialized tools to handle multi-modal documents. The ``read" tool takes a page ID and a sub-query, using a Vision-Language Model (VLM) to extract and interpret relevant text, charts, and images from the specified page. This enhances our framework's ability to process visual information, enabling more accurate understanding and reasoning of multimodal content within documents. The ``search" tool uses BM25 with a sub-query to perform fast, keyword-based retrieval across the document, returning the Top-K most relevant text snippets. BM25 is chosen for its efficiency and effectiveness in rapid information lookup.
\subsubsection{Self-Refine \& Information Seeking}
This phase forms the core iterative loop of MM-Doc-R1. The agent dynamically decomposes the complex question into executable \textbf{sub-queries} and, in each iteration, selects and invokes appropriate \textbf{Tools} (read or search) to retrieve needed information. It generates a sub-query, chooses a tool, executes it with parameters, and analyzes the output. If the sub-query remains unresolved, the agent refines its plan and iterates. This process enables adaptive, step-by-step reasoning through complex document queries.

\subsubsection{Answer Generation}
After gathering sufficient information, the agent enters the \textbf{answer generation} phase. It synthesizes all retrieved context---including tool outputs and OCR text---enabling the LLM to reason over the evidence and produce a final, accurate, and coherent answer that fully addresses the original query.

\subsection{Training Method: Similarity-based Policy Optimization (SPO)}
To enable our agent to learn optimal strategies for tool invocation, sub-query decomposition, and overall workflow navigation, we propose a novel reinforcement learning algorithm called Similarity-based Policy Optimization (SPO). SPO serves as a significant enhancement to existing policy optimization techniques, particularly improving upon the GRPO algorithm by providing a more precise and stable learning signal for agentic decision-making. Figure~\ref{fig:algorithm1} shows the difference between SPO and GRPO.

\subsubsection{Group Relative Policy Optimization}

GRPO is a popular algorithm proposed by DeepSeek; it is an improvement over PPO. The loss function of GRPO is


\begin{equation}
\begin{aligned}
\mathcal{J}_{\text{GRPO}}(\theta) = \mathbb{E}_{q, o_{1:G}} \biggl[ & \frac{1}{G} \sum_{i=1}^{G} \frac{1}{|o_i|} \sum_{t=1}^{|o_i|} \min \Bigl( r_t(\theta) \hat{A}_{i,t}, \\
& \operatorname{clip}(r_t(\theta), 1-\varepsilon, 1+\varepsilon) \hat{A}_{i,t} \Bigr) \\
& - \beta \mathbb{D}_{\text{KL}} [ \pi_\theta \| \pi_{\text{ref}} ] \biggr]
\end{aligned}
\label{eq:grpo}
\end{equation}
where $r_t(\theta) = \frac{\pi_\theta(o_{i,t}|q, o_{i,<t})}{\pi_{\theta_{\text{old}}}(o_{i,t}|q, o_{i,<t})}$ is the importance sampling ratio, $\hat{A}_{i,t}$ is the advantage estimate, $\varepsilon$ is the clipping threshold, and $\beta$ controls the KL regularization strength.

In multi-turn RL, GRPO calculates the advantage by comparing the reward of the current generated policy ($T_i$) with the average reward of all policies within the batch. The advantage function for GRPO is given by:

\begin{equation}
A_{\text{GRPO}}(T_i) = R(T_i) - \frac{1}{N} \sum_{j=1}^{N} R(T_j)
\label{eq:grpo-advantage}
\end{equation}

Here, $R(T_i)$ represents the reward obtained for the current trajectory $T_i$, and $N$ is the total number of trajectory in the current group. Note that in a multi-turn reinforcement learning training process, the entire response in the same trajectory receives the same reward. 

In traditional single-turn or fixed-environment RL settings, it's typically assumed that all responses within a group share the same initial conditions, often referred to as the ``prompt". However, this fundamental assumption becomes problematic in multi-turn RL training. As training progresses across multiple interaction rounds, the diversity among individual trajectories rapidly increases. This divergence means that even within the same group, two trajectories can evolve under significantly different intermediate states or contexts. Consequently, one cannot directly assume that the rewards derived from these disparate trajectories are directly comparable, as their underlying conditions are no longer uniform.

\subsubsection{Similarity-based Policy Optimization}

In multi-turn reinforcement learning, a trajectory is typically modeled as a sequence of states and actions:
\begin{equation}
    s_0 \rightarrow a_0 \rightarrow s_1 \rightarrow a_1 \rightarrow \cdots \rightarrow s_T \rightarrow a_T,
\end{equation}
where $ s_t $ denotes the state (eg. prompt and the environment feedback) at step $ t $, and $ a_t $ is the agent's response. In GRPO, $ n $ trajectories are sampled in parallel from the same initial state $ s_0 $, and a shared baseline $ V(s_0) $ is used for advantage estimation. While computationally efficient, this approach assumes that all trajectories remain semantically aligned throughout the interaction, which is an assumption that quickly breaks down as responses diverge over turns.

As dialogue progresses, even trajectories starting from the same prompt can evolve into significantly different contexts due to stochastic generation and feedback dynamics. Consequently, their value estimates $ V(s_t) $ become increasingly heterogeneous. Using a single baseline derived from $ s_0 $ introduces high bias in advantage estimation, especially for later turns, leading to unstable policy updates.

To address this, we propose Similarity-based Policy Optimization (SPO), which replaces the uniform baseline with a semantically weighted average over rewards in the batch. The key insight is that trajectories with similar semantic content are more likely to share underlying value structures and thus should serve as better baselines for one another.

The advantage in SPO is defined as:

\begin{equation}
A_{\text{SPO}}(T_i) = R(T_i) - \sum_{j=1}^{N} w_{ij} R(T_j),
\end{equation}
where $ R(T_i) $ represents the total reward of trajectory $ T_i $. The weight $ w_{ij} $ reflects the semantic similarity between trajectory $ T_i $ and $ T_j $, and is normalized across the group:

\begin{equation}
\begin{split}
w_{ij} = \frac{\text{similarity}(\text{emb}(T_i), \text{emb}(T_j))}
{\sum_{k=1}^{N} \text{similarity}(\text{emb}(T_i), \text{emb}(T_k))},
\end{split}
\end{equation}
Here, $\text{emb}(T)$ denotes the dense vector representation of trajectory $ T $, computed using the BGE-M3 model~\cite{chen2024bge}, which is frozen during training. The $\text{similarity}$ function computes the cosine similarity between embeddings.

By constructing a dynamic, content-aware baseline, SPO reduces estimation variance and mitigates bias caused by trajectory divergence. It effectively prioritizes comparisons within semantically coherent groups, yielding more accurate advantages and stabler learning, particularly in long-horizon, multi-turn settings where traditional baselines fail.
\subsubsection{Reward Function Design}
We employ a comprehensive set of metrics to evaluate the performance of all models, reflecting both answer accuracy and the ability to correctly identify unanswerable questions. Our primary reward signal for reinforcement learning is the \textbf{Final Reward}, calculated by summing the \textbf{read page Recall} and a \textbf{Correctness Score} for the final answer. The Final Reward is defined as:
\[
\text{Final Reward} = \text{Recall} + \text{Correctness Score}
\]
The read page Recall measures how effectively the system directs the agent to relevant pages containing the answer, defined as:
\[
\text{Recall} = \frac{|\text{set(read pages)} \cap \text{set(evidence pages)}|}{|\text{set(evidence pages)}|}
\]
The Correctness Score for the final answer is determined by following the answer judgment methodology from MMlongbench-doc~\cite{ma2024mmlongbench}, specifically, we leverage \textbf{Qwen2.5-72B-Instruct} to extract a precise answer, and then perform a matching calculation against the ground-truth answer to derive this score. 

\section{Experiments}

\begin{table*}[t]
\centering
\small
\setlength{\tabcolsep}{5.5pt}
\renewcommand{\arraystretch}{0.9}
\begin{tabular}{@{}l*{10}{c}@{}}
\toprule
 & \multicolumn{5}{c}{\textbf{Evidence Modality}} & \multicolumn{3}{c}{\textbf{Evidence Count}} & \multicolumn{2}{c}{\textbf{Overall}} \\
\cmidrule(lr){2-6}\cmidrule(lr){7-9}\cmidrule(lr){10-11}
\textbf{Method} & Text & Layout & Chart & Table & Figure & Single & Multi & Unans. & ACC & F1 \\
\midrule
\multicolumn{11}{c}{\textit{Human Performance}} \\
\cmidrule(lr){1-11}
Human & --- & --- & --- & --- & --- & --- & --- & --- & 65.8 & 66.0 \\
\midrule
\multicolumn{11}{c}{\textit{Upper Bounds (Ground-Truth Evidence)}} \\
\cmidrule(lr){1-11}
Qwen2.5-VL-7B & 33.8 & 38.7 & 31.8 & 32.3 & 34.1 & 46.6 & 20.5 & 92.8 & 46.8 & 41.7 \\
Qwen3-8B & 44.3 & 37.7 & 25.7 & 59.3 & 22.8 & 42.7 & 35.7 & 89.2 & 49.6 & 46.9 \\
\midrule
\multicolumn{11}{c}{\textit{RAG Baselines}} \\
\cmidrule(lr){1-11}
BM25 & 30.9 & 23.4 & 22.3 & 28.5 & 9.2 & 30.7 & 14.1 & 88.3 & 36.4 & 31.0 \\
BGE-M3 & 32.0 & 20.8 & 21.7 & 40.3 & 14.7 & 35.4 & 18.5 & 84.3 & 39.3 & 34.8 \\
Colqwen & 27.8 & 25.0 & 16.5 & 22.4 & 23.7 & 33.9 & 13.7 & 82.5 & 36.5 & 31.2 \\
Mdoc agent & 33.1 & 29.3 & 25.8 & 32.6 & 30.0 & 43.7 & 18.4 & 43.4 & 35.0 & 33.3 \\
M3doc RAG & 39.2 & 26.7 & 29.8 & 39.0 & 32.0 & 50.3 & 21.2 & 40.7 & 38.4 & 36.7 \\
\midrule
\multicolumn{11}{c}{\textit{Ours: MM-Doc-R1}} \\
\cmidrule(lr){1-11}
Qwen3-4B & 28.9 & 23.8 & 23.1 & 35.3 & 22.4 & 37.1 & 18.6 & 72.2 & 37.7 & 32.2 \\
\quad +GRPO & 36.3 & 35.2 & 29.5 & 40.2 & 27.1 & 44.5 & 22.5 & 58.7 & 39.9 & 36.3 \\
\quad +SPO & 41.1 & 37.2 & 35.6 & 47.2 & 30.5 & 50.5 & 27.5 & 68.0 & 46.0 & 41.2 \\
Qwen3-8B & 39.6 & 37.6 & 37.8 & 45.6 & 27.3 & 47.3 & 27.5 & 73.9 & 45.7 & 41.5 \\
\quad +GRPO & 40.9 & 36.8 & 35.7 & 49.9 & 28.3 & 48.5 & 30.2 & 60.5 & 44.7 & 41.9 \\
\quad +SPO & \textbf{46.2} & \textbf{38.1} & \textbf{40.8} & \textbf{52.8} & \textbf{35.9} & \textbf{56.0} & \textbf{31.2} & 68.2 & \textbf{49.7} & \textbf{46.1} \\
\bottomrule
\end{tabular}

\caption{Performance comparison on the MMLongBench-Doc dataset (1,082 questions).
The evaluation includes text-based RAG methods (BM25, BGE-M3), multi-modal RAG (ColQwen2.5-7B-multilingual), and agent-based systems (Mdoc Agent, M3doc RAG). Text-based methods use the top-4 retrieved pages and Qwen3-8B for answer generation. Multi-modal RAG uses the top-4 retrieved image pages and Qwen2.5-VL-7B. Agent-based methods operate over the top-5 retrieved pages, we use Qwen2.5-VL-7B as the VLM and Qwen3-8B as the LLM. 
Metrics include overall Accuracy (ACC) and F1, as well as performance on sub-categories by evidence modality (Text, Layout, Chart, Table, Figure), number of required evidences (Single, Multi), and unanswerable questions. 
Best scores among baselines and our methods are marked in bold, second-best in underline, considering only ``RAG Baselines", and ``Ours MM-Doc-R1" sections.}
\label{tab:main_results}
\end{table*}
This section details the experimental setup, evaluation metrics, and comprehensive results demonstrating the efficacy of our proposed \textbf{MM-Doc-R1} framework, particularly highlighting the performance gains achieved through our RL algorithm \textbf{SPO}.

\subsection{Experimental Setup}
The MMLongbench-Doc dataset serves as our primary benchmark for evaluating multi-modal long document question answering. This dataset features complex documents requiring multi-step reasoning and spans various content types. For RL training, we use a subset of 300 samples from LongDocURL as the validation set, and the remaining data as the training set.

\begin{figure*}[t!] 
    \centering 
    \includegraphics[width=1.0\textwidth]{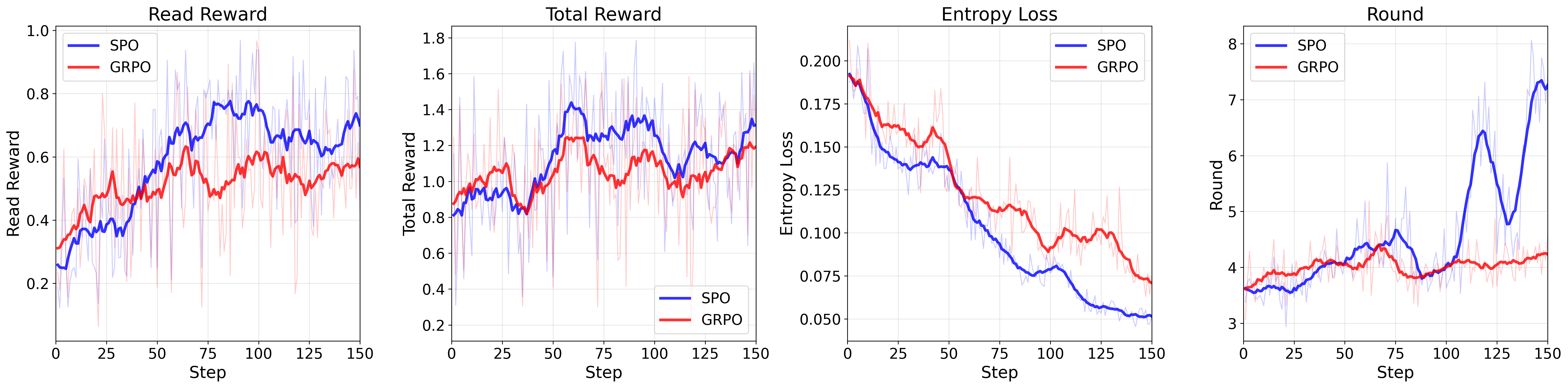} 
    \caption{Comparison of SPO and GRPO in Training; the base model in this figure is Qwen3-8B.} 
    \label{fig:rl_comparison_chart} 
\end{figure*}

\subsection{Results and Discussion}
Our experimental results, summarized in Table~\ref{tab:main_results}, clearly demonstrate the superior performance of MM-Doc-R1, particularly when trained with SPO, across various dimensions of the MMLongBench-Doc benchmark. As shown in the table, even in its untrained form, MM-Doc-R1 with Qwen3-8B achieves an overall Accuracy (ACC) that surpasses the current state-of-the-art baseline by 10.4\%. In single-evidence questions---where the answer can be derived from a single retrieved page---our method outperforms the best existing approach by 5.7\%. More notably, on multi-evidence questions that require information integration across multiple pages and often involve multi-hop reasoning, MM-Doc-R1 achieves a gain of 10.0\%, highlighting its strong capability in handling complex, long-context reasoning tasks. Furthermore, our framework consistently achieves top performance across all evidence modalities, including Text, Layout, Chart, Table, and Figure, demonstrating its robustness and effectiveness in processing heterogeneous multi-modal document content.

In terms of reinforcement learning, SPO exhibits clear advantages over GRPO. When applied to the Qwen3-4B model, SPO improves overall accuracy by 6.1\% compared to GRPO; on Qwen3-8B, the improvement reaches 5.0\%. This consistent gain confirms the effectiveness of our semantic similarity-based advantage estimation in mitigating the bias introduced by trajectory divergence in multi-turn dialogue settings. As illustrated in Figure~\ref{fig:rl_comparison_chart}, SPO not only achieves higher final performance but also demonstrates more stable and faster convergence during training. These results validate our hypothesis that leveraging semantically aligned trajectories as dynamic baselines leads to more accurate policy updates, especially in long-horizon, multi-step reasoning scenarios.
\subsection{Ablation Study}
\begin{table}[t]
\small
\centering
\begin{tabular}{lccc}
\toprule
Model & ACC & F1 & Unanswerable \\
\midrule
MM-Doc-R1 (Qwen3-8B) & \textbf{45.7} & \textbf{41.5} & 73.9 \\
\midrule
\hspace{1em} w/o TOC & 44.3 & 39.9 & 73.5 \\
\hspace{1em} w/o Read pages' OCR & 43.4 & 38.7 & 74.7 \\
\hspace{1em} w/o VLM Read & 42.1 & 37.3 & \textbf{81.6} \\
\bottomrule
\end{tabular}
\caption{Ablation study of MM-Doc-R1 components on MMLongBench-Doc, using Qwen3-8B. All metrics are in \%.}
\label{tab:ablation_study}
\end{table}
In order to assess the individual contributions of each component within MM-Doc-R1, we conducted an ablation study, summarized in Table~\ref{tab:ablation_study}. The removal of any component consistently led to a degradation in performance, underscoring the vital role of the table of contents (TOC), page OCR reading, and VLM reading modules in enhancing the model's overall efficacy. These findings collectively emphasize the synergistic effectiveness of all components within MM-Doc-R1.

\subsection{Recall Analysis}

Table~\ref{tab:recall_top5} presents the Recall performance of various models. Our MM-Doc-R1 framework achieves a recall of 66.3\% when just read 3.35 pages, significantly outperforming traditional BM25 (42.0\%), embedding-based BGE-M3 (51.3\%), and multi-modal Colqwen-2.5-VL-7B (65.4\%). Notably, MM-Doc-R1 achieves this highest recall while requiring an average of just 3.35 pages read, underscoring the effectiveness of its integrated agentic framework. This superior recall ensures the agent accesses more relevant evidence, crucial for accurate question answering in multi-modal contexts.
\begin{table}[htbp]
    \centering
    \begin{tabular}{lcc} 
        \toprule 
        Model & average pages & Recall  \\
        \midrule 
        BM25 & 5 & 42.0  \\
        BGE-M3 & 5 & 51.3  \\
        Colqwen-2.5-VL-7B & 5 & 65.4  \\

        \midrule 
        MM-Doc-R1(SPO) & 3.35 & 66.3  \\
        \bottomrule 
    \end{tabular}
    \caption{Recall Performance} 
    \label{tab:recall_top5} 
\end{table}


\section{Conclusion}
We presented \textbf{MM-Doc-R1}, addressing the limitations of single-pass RAG in long-document visual QA through an iterative, agentic workflow. Central to our framework is \textbf{Similarity-based Policy Optimization (SPO)}, which mitigates baseline estimation bias in multi-turn RL by leveraging semantic trajectory similarity for precise reward averaging. Our experiments on MMLongbench-Doc demonstrate that MM-Doc-R1 outperforms prior baselines by 10.4\%, with SPO significantly surpassing standard GRPO across multiple model scales. These results validate that integrating vision-aware reasoning with trajectory-informed RL effectively advances the state-of-the-art for complex, multimodal information discovery.

\section{Limitations}

While MM-Doc-R1 and the SPO algorithm demonstrate substantial improvements in long-document visual reasoning, several inherent limitations should be noted.

First, the effectiveness of our framework is \textbf{partially dependent on the quality of initial document parsing}. Although MM-Doc-R1 employs a robust seeker to navigate content, its planning and retrieval efficiency still rely on the fidelity of the structural metadata (such as the Table of Contents) and the OCR accuracy of the ingestion engine. In scenarios where documents are severely degraded or lack standard hierarchical formatting, the performance may be constrained by the noise introduced during the pre-processing stage.

Second, our current study primarily focuses on \textbf{understanding static documents}. The proposed multi-turn workflow and reinforcement learning strategy are optimized for fixed document formats like PDFs and high-resolution images. However, real-world digital documents can be dynamic or semi-structured (e.g., interactive reports or web-based content). The adaptation of the agentic seeker to environments where document content or layouts might dynamically evolve during interaction remains an area for future exploration.

Finally, while SPO effectively reduces baseline estimation bias, its current validation is centered on English-centric benchmarks. The cross-lingual robustness and generalizability of the seeker across diverse linguistic structures have yet to be extensively investigated.

\section*{Acknowledgments}
The authors wish to thank the anonymous reviewers for their helpful comments. This work was partially funded by National Key R\&D Program of China No.2025ZD0215702, National Natural Science Foundation of China (No. 62576106, 62476061, 62376061).

\bibliography{custom}

\clearpage
\appendix

\section{Appendix}
\label{sec:appendix}
\subsection{Evaluation Metrics}
We adopt the exact same evaluation protocol as MMLongBench-Doc~\cite{ma2024mmlongbench}. We report the overall F1 score (F1) and the overall accuracy (ACC). To assess modality-specific performance, we break down accuracy by content type ----Text, Layout, Chart, Table, and Figure ----reflecting the diverse modalities present in long documents. Questions are further categorized by the number of evidences required: \textbf{Single Evidence} and \textbf{Multi Evidence}, to evaluate reasoning capabilities in simple versus complex scenarios. Additionally, we measure \textbf{Unanswerable Accuracy}, i.e., the percentage of questions correctly identified as unanswerable, which is crucial for assessing the robustness of QA systems in real-world settings.
\subsection{Baselines}
To rigorously evaluate \textbf{MM-Doc-R1}, we compare it against a comprehensive set of state-of-the-art RAG and agent-based baselines across different paradigms. All text-based methods use Qwen3-8B to generate answer, and multi-modal methods use Qwen2.5-VL-7B.

For text-based RAG methods, we include BM25 and BGE-M3. BM25 is a classical sparse retrieval approach using keyword matching. BGE-M3 is a dense retrieval method that leverages semantic embeddings for document retrieval. Since the original PDFs are image-based, we apply Doc2X for high-fidelity OCR to extract textual content before indexing. The top-4 retrieved pages are used as evidence for answer generation.

For multi-modal RAG, we evaluate 
ColQwen2.5-7b-multilingual, which retrieves relevant document pages using vision-language understanding and performs end-to-end answer generation from images. This method uses the top-4 retrieved pages as input to maximize coverage of visual and structural content.

We also compare against recent agent-based document QA systems: Mdoc Agent~\cite{han2025mdocagent} and M3doc RAG~\cite{cho2024m3docrag}. Both systems operate over the top-5 retrieved pages to ensure consistent input scope.

All methods use the same candidate evidence pool and are evaluated under consistent metrics to ensure fair comparison.

\subsection{Implementation Details}
Our framework is implemented using the Qwen3-8B and Qwen3-4B Large Language Models as the core agent orchestrator. The read tool incorporates Qwen2.5-7B-VL, a Vision-Language Model (VLM) designed for multi-modal content extraction, which is employed in a zero-shot manner (i.e., without additional training). The search tool utilizes the BM25 algorithm for text retrieval, returning the top 10 most relevant pages for each query. The final input to the answer model consists of a short snippet from the search result and the detailed context from the read result, where the search content occupies only a minimal portion of the overall context and is not fed in full. For reinforcement learning training, we specifically evaluate two policy optimization algorithms: GRPO and our proposed SPO. Both GRPO and SPO are trained under the same settings, except for the advantage estimation method.

We employ the \textbf{verl} framework as our reinforcement learning (RL) backbone. Within this framework, we apply a custom patch to the advantage computation function to accommodate our proposed algorithmic enhancements. 

For training infrastructure, we utilize a distributed setup powered by \textbf{8 NVIDIA H100 GPUs}. The system leverages a hybrid of data and model parallelism to support the efficient training of large-scale policy models (e.g., Transformer-based architectures). All computations are accelerated via CUDA, and Automatic Mixed Precision (AMP) is employed to significantly enhance training throughput while maintaining numerical stability.

Our implementation is built upon the \textbf{GRPO}  algorithm, with key modifications to the advantage computation and update dynamics. The primary hyperparameters are configured as follows:
    
- \textbf{Rollout length}: Set to \texttt{4}, meaning 4 steps of environment interaction are collected per policy sampling cycle before performing value estimation and policy update. This balances training stability with timely feedback.
    
- \textbf{Batch size}: Set to \texttt{4}, indicating the number of complete rollout trajectories used in each policy update. Given that each rollout contains multiple time steps, the effective number of state-action pairs per update is $4 \times 4 = 16$.
    
- \textbf{Temperature}: Set to \texttt{0.8}, which controls the level of exploration in the policy distribution. A lower temperature encourages more deterministic outputs, promoting convergence on high-confidence actions while preserving moderate exploration.
    
- \textbf{Learning rate}: The policy network is trained with a learning rate of \texttt{5e-7} ($5 \times 10^{-7}$), using the Adam optimizer. This conservative learning rate is chosen to accommodate the high-dimensional parameter space and high-precision gradient computation enabled by the H100 GPUs, helping to prevent policy collapse.
    
- \textbf{KL coefficient}: Set to \texttt{0.005}, serving as a regularization term to constrain the Kullback--Leibler (KL) divergence between the old and new policies during updates. This prevents excessive policy shifts and enhances training robustness.

\subsection{Computational Overhead Analysis}
\label{sec:overhead}

Since the iterative agentic workflow naturally incurs more LLM/VLM calls than single-pass RAG, we quantify the actual per-query latency to assess practical usability. Table~\ref{tab:latency} reports the wall-clock time per question on the full MMLongBench-Doc set (1,082 questions).

\begin{table}[h]
\small
\centering
\begin{tabular}{lccc}
\toprule
Method & Avg & Min & Max \\
\midrule
ColQwen2.5 RAG (one-pass) & $\sim$40 & -- & -- \\
MM-Doc-R1 (Qwen3-4B, no train) & 84.47 & 10 & 724 \\
MM-Doc-R1 (Qwen3-8B, no train) & 109.32 & 11 & 397 \\
MM-Doc-R1 (Qwen3-8B, SPO)      & 154.44 & 23 & 316 \\
\bottomrule
\end{tabular}
\caption{Per-question latency (seconds) on MMLongBench-Doc.}
\label{tab:latency}
\end{table}

\paragraph{Test-time scaling, not free overhead.} MM-Doc-R1 belongs to the family of \textit{test-time scaling} approaches (cf.\ DeepSeek-R1, Gemini Deep Research): additional compute is exchanged for substantially higher accuracy on multi-hop, multi-evidence questions. The 109.32~s average for the 8B variant is well within practical document-analysis usage, while delivering the +10.4\% absolute ACC gain reported in §4.

\paragraph{Lightweight retrieval scales better with document length.} Embedding-based RAG must vectorize \emph{all} pages of every new document, so its latency grows roughly linearly with document length and incurs additional storage cost. MM-Doc-R1 instead uses BM25 for first-stage retrieval, which requires no precomputed embeddings and supports real-time DocQA on freshly arrived documents. Consequently, as documents grow longer, the relative overhead of MM-Doc-R1 over RAG \emph{shrinks} rather than grows, making the agentic workflow particularly suited to long-document scenarios.

\paragraph{Why the SPO-trained model is slower.} The SPO-trained agent (154.44~s) takes longer than the untrained one (109.32~s) because RL training encourages more proactive multi-turn exploration, which directly contributes to the recall and accuracy gains; the cost--benefit trade-off remains favorable.

\subsection{BM25 Topk choose}
\begin{figure}[h] 
    \includegraphics[width=0.45\textwidth]{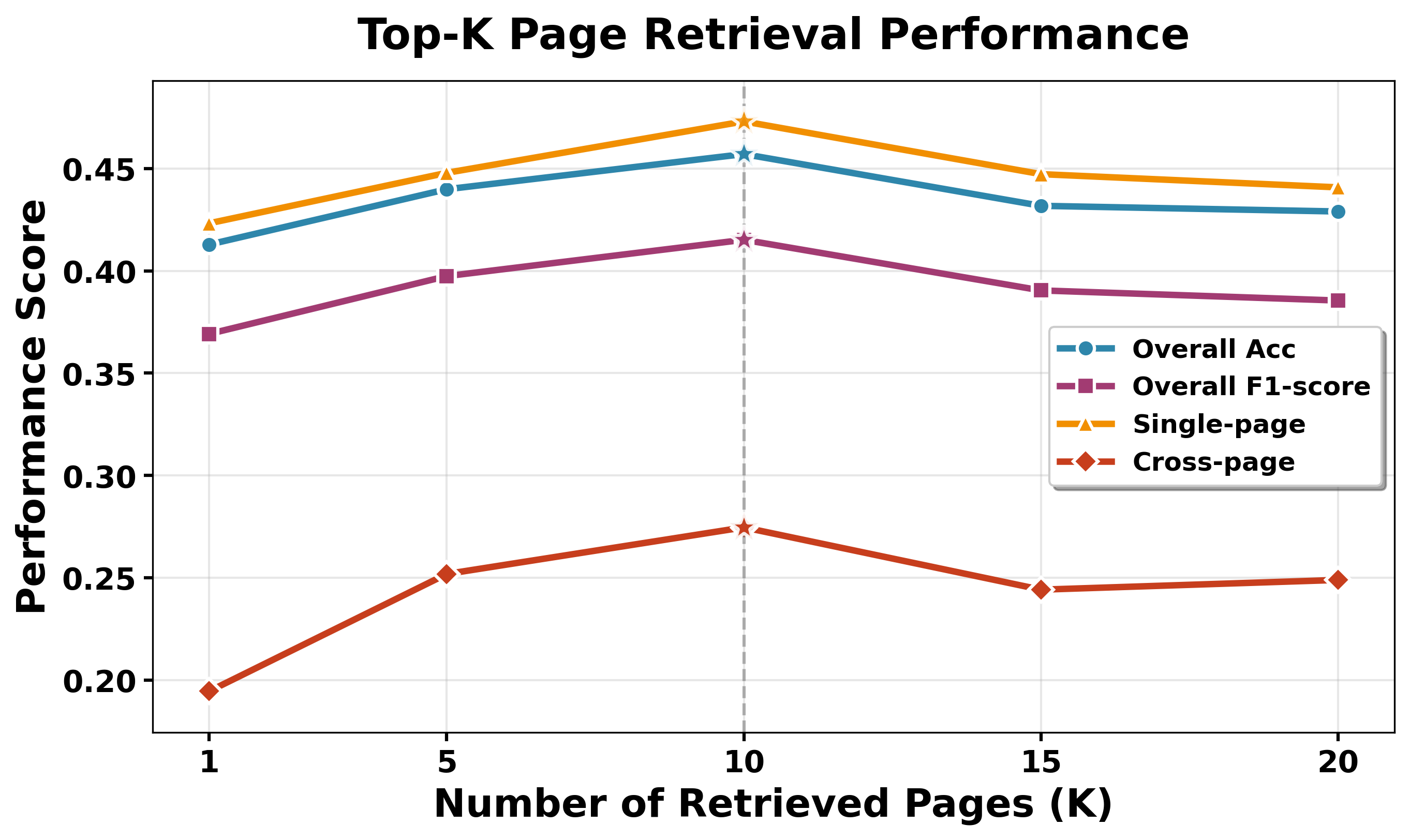}
    \caption{Performance with BM25 topk}
    \label{fig:topk_compare}
\end{figure}
Figure ~\ref{fig:topk_compare} presents the impact of varying the top-$k$ parameter on the performance of MM-Doc-R1.Performance across all metrics generally improves with increasing values of parameter $k$. However, substantial gains are primarily observed when $k < 10$. For $k > 10$, the rate of improvement slows, and performance may even decline, attributable to the increased context length potentially diluting relevant information. Consequently, $k=10$ was selected as our experimental setting.

\subsection{Prompt}
\begin{lstlisting}[caption={Prompt for Planner}]
You are an expert in **long document analysis** and **information retrieval planning**, capable of designing systematic and efficient exploration strategies using multiple types of clues.

Your task:
Based on the user's query, the document's table of contents (TOC) and the document's list of figures, create an **initial research plan** that is goal-oriented and progresses step by step, helping downstream modules efficiently understand and extract information from the document.

You may use the following tools in your planning:
(1) Search Tool: Input keywords to search for relevant content in the document. It returns page IDs that contain those keywords. You can search multiple keywords at once.
(2) Read Page: Read up to 5 pages of the document and prepare the content for analysis.

Notes:
- Your plan should not exceed 10 steps. Keep the logic clear and progressive.


Please output the initial plan in the following format:
<output_format>
Here is my proposed initial plan.
(1) ...
(2) ...
(3) ...
(4) ...
(5) ...
(6) ...
...
</output_format>

{example}

<input>
- Query: {query}
- document'd table of contents:
<document_toc>
{document_toc}
</document_toc>
<list_of_figures>
{list_of_figures}
</list_of_figures>
</input>
\end{lstlisting}

\begin{lstlisting}[caption={Prompt for Seeker}]
    You are performing a step-by-step task of information extraction and understanding. Based on the current query goal and the steps already taken (plan_done), you need to:

(1) First, explain your reasoning process, including:
- What information is still missing or unclear?
- What is the next key issue or sub-goal to address?
- Which tools can help fill in this gap? Do you need a combination of text and visual content?

(2) Then, based on the above analysis, decide on the next tool invocation(s).

You can use the following tools. In each step, you may choose a reasonable number of queries:

---

**(1) Search Tool**: Search whether certain keywords or topics appear in the document. Returns brief summaries of the pages where matches are found.  
Format:
```tool_code
<search>["keyword1", "keyword2"]</search>
````

**(2) Read Page**: Read the content of specified pages. Returns a summary relevant to your query.
You may read up to 3 pages at a time. Clearly state your intent for using this tool.
Format:

```tool_code
<read> 
[{{
    "page_ids": [4, 5, 8],
    "query": "sub_query1"
}}, {{
    "page_ids": [13, 19, 20],
    "query": "sub_query2"
}}]
</read>
```

**(3) Termination Marker**: When you determine that enough information has been gathered and the task can end, return:

```tool_code
<FINISH>
```

Special Notes:

* Each step must be concise, strategic, and limited to one tool only.
* You are encouraged to demonstrate **structured, progressive strategic thinking**.

{example}

Query:
{query}

Steps already taken:
{plan_done}

Your reasoning and next-step plan:
\end{lstlisting}

\begin{lstlisting}[caption={Prompt for reader}]
You are a professional page summary expert. Your task is to extract key information about the origin query and sub query from given pages.

Your input consists of:
1. An origin query
2. A sub query

#### Instructions
1. First, extract all visual elements:
   - Tables
   - Figures
   - Charts
   - Images
   - Text content

2. Then, identify information relevant to:
   - Origin Query
   - Sub Query

3. Format your output as:
   - Visual Elements Summary
   - Query-Relevant Information (text and visual elements)
   - Key Findings

#### Important Notes
- Base your analysis strictly on the provided images
- Do not make assumptions or add information beyond what is shown
- If required information is missing, clearly state: "Cannot answer due to insufficient data"
- The sub query is the most recent and relevant query, while the origin query is the earlier context query

Input:

Origin Query:
{origin_query}

Sub Query:
{sub_query}

Image of Pages:
\end{lstlisting}

\begin{lstlisting}[caption={Prompt for answer}]
Please answer the question using only the available information. Do not fabricate or assume any details beyond what has been provided.
If the necessary information is not available, clearly state that you cannot answer the question due to lack of relevant data.
question:{origin_query}
Related Information:{past_information}
\end{lstlisting}

\subsection{Use of AI}

We utilized generative AI tools to assist in refining the manuscript's language and optimizing the structure of our source code. We used these tools to improve clarity and coding efficiency; however, we reviewed and edited all outputs to ensure technical accuracy. We take full responsibility for the final content of the manuscript and the integrity of the implemented code.

\end{document}